# A Global Information Based Adaptive Threshold for Grouping Large Scale Optimization Problems

An Chen[1], Yipeng Zhang
Autocontrol Institute, Xi'an Jiaotong University
Xi'an, Shaanxi, 710049, P.R. China
chenan123@stu.xjtu.edu.cn[1]

Zhigang Ren
Autocontrol Institute, Xi'an Jiaotong University
Xi'an, Shaanxi, 710049, P.R. China
renzg@mail.xjtu.edu.cn

Yongsheng Liang
Autocontrol Institute, Xi'an Jiaotong University
Xi'an, Shaanxi, 710049, P.R. China
liangyongsheng@stu.xjtu.edu.cn

Bei Pang
Autocontrol Institute, Xi'an Jiaotong University
Xi'an, Shaanxi, 710049, P.R. China
beibei@stu.xjtu.edu.cn

## ABSTRACT

By taking the idea of divide-and-conquer, cooperative coevolution (CC) provides a powerful architecture for large scale global optimization (LSGO) problems, but its efficiency relies highly on the decomposition strategy. It has been shown that differential grouping (DG) performs well on decomposing LSGO problems by effectively detecting the interaction among decision variables. However, its decomposition accuracy depends highly on the threshold. To improve the decomposition accuracy of DG, a global information based adaptive threshold setting algorithm (GIAT) is proposed in this paper. On the one hand, by reducing the sensitivity of the indicator in DG to the roundoff error and the magnitude of contribution weight of subcomponent, we proposed a new indicator for two variables which is much more sensitive to their interaction. On the other hand, instead of setting the threshold only based on one pair of variables, the threshold is generated from the interaction information for all pair of variables. By conducting the experiments on two sets of LSGO benchmark functions, the correctness and robustness of this new indicator and GIAT were verified.

## 1 INTRODUCTION

Nowadays, the dimensions of optimization problems involved in scientific research and engineering practice are getting higher and higher [1], so how to solve the large scale global optimization (LSGO) problems becomes more and more important and it has become an active research field over the last decade. For the black-box characteristics of LSGO, the gradient-free evolutionary algorithms (EAs) are the major approaches being taken to resolve them. However, due to "the curse of dimensionality" that the size of search space grows exponentially with respect to the number of decision variables, conventional EAs cannot adequately explore the whole space of LSGO problems within acceptable computation times [2, 3].

To alleviate this limitation, a special architecture named cooperative coevolution (CC) was proposed [4]. By taking the strategy of "divide-and-conquer", CC provides a natural but also efficient approach for solving LSGO problems. It first decomposes the original LSGO problem into a certain number of lower dimensional sub-problems and then cooperatively optimizes these sub-problems with a conventional EA. From the framework of CC, it can be concluded that the decomposition strategy plays a vital role. An ideal decomposition algorithm should achieve the aim that separable variables could be grouped into different sub-problems and all nonseparable variables could be grouped into the same one. After decomposition process, the interaction between sub-problems should be minimized. Besides, for the limited times of objective fitness evaluations (FEs) supplied, the whole decomposition process should be accomplished within as fewer FEs as possible.

In recent years, great efforts have been put on decomposition process and many decomposition algorithms were proposed [5]. According to their core ideas, the existing algorithms can be classified into three categories, including static decomposition strategy, dynamic decomposition strategy and learning-based decomposition strategy. Specifically, static strategy [3, 4, 6] is the initial decomposition algorithm where the original LSGO problems are decomposed in a fixed pattern and each sub-problem is kept unchanged during the whole CC process.

Dynamic strategy [7, 8] is different from the former, it first randomly assigns all variables to a certain number of groups and then regroups them before every evolution cycle. The two strategies perform well on separable problems, but for the reason that both of them do not take the interactions among variables into account, their performances degenerate quickly on the problems with nonseparable variables. [9] To make near-optimal decomposition for all kinds of functions, learning-based strategy explores an entirely different path. It groups the variables into different subcomponents based on the detected interdependency information.

Different grouping (DG) [9] is a typical learning-based decomposition algorithm which can identify the interactions among variables effectively, and it has shown superior performance as compared with other algorithms. However, DG can only detect the direct interdependency information and it consumes lots of FEs on some problems. Aiming at these shortcomings, some new versions of DG algorithms were proposed and achieved pretty good results [10-14]. But there is still a problem for DG that needs to be improved. The decomposition accuracy of DG relies highly on the threshold parameter which is used to compensate for the computational roundoff error. But the threshold is very sensitive to the roundoff error and the magnitude of contribution weights of different subcomponents, so it is very difficult to find a suitable one for some problems. Up to now, the existing threshold setting methods are all unsatisfactory [9, 11, 13], and the decomposition accuracy of DG is limited by the threshold parameter.

To address the above problem and improve the decomposition accuracy of DG-based algorithms, this paper proposed a global information based adaptive threshold setting algorithm (GIAT). In GIAT, a new indicator is developed for each pair of variables firstly. This indicator is much more sensitive to the interaction between variables and less sensitive to other factors, such as the roundoff error and the magnitude of contribution weights of different subcomponent. In other words, there will be great difference only between the indicator for two nonseparable variables and the indicator for two separable variables. Therefore, when we sort the indicators for each pair of variables in ascending order, the two indicators with the greatest difference are the biggest indicator for two separable variables and the smallest indicator for two nonseparable variables, respectively. Then a suitable threshold parameter can be easily set based on these two indicators. The correctness and robustness of this new indicator and GIAT are verified on the functions in both CEC2010 and CEC2013 benchmark [15, 16] suites in the final.

The remainder of this paper is organized as follows: Section 2 reviews the related work, including some typical decomposition algorithms and threshold setting methods for DG-based algorithms. Section 3 gives the detailed description of the new indicator and GIAT. Section 4 presents the experimental results and analyses. Finally, the conclusions and future work are discussed in Section 5.

## 2 RELATED WORK

In this section, we first review some typical decomposition algorithms, and then a brief introduction for the existing threshold setting methods for DG-based algorithms is given.

### 2.1 Decomposition Algorithms

Decomposition plays a vital role in ensuring the performance of CC, and by now a variety of decomposition algorithms have been developed. They can be generally divided into three categories, including static decomposition strategy, dynamic decomposition strategy, and learning-based decomposition strategy. In this part, we will review them in sequence.

Static strategy is the simplest strategy. In its simplest form, an $n$-dimensional problem is partitioned into $k$ $s$-dimensional sub-problems where $n = k \cdot s$. For different static algorithms, $k$ is set to different values, such as one [4], two [6] or any random number [3]. But for its neglect of variable interaction, static strategy only performs well in fully separable functions [9]. To remedy this defect, dynamic strategy was developed and random grouping [7] and MLCC [8] are two typical dynamic algorithms. Random grouping randomly allocates all the decision variables into $k$ $s$-dimensional sub-problems before every co-evolution cycle instead of using a static grouping. To tackle the issue that it is difficult to specify a value for $k$, MLCC is proposed. It has been shown [5, 10] that the performance of dynamic strategy is better than that of static strategy. However, it can only correctly decompose the problems with just a few nonseparable variables. Once the number of nonseparable variables increases, its performance gets worsened.

In order to alleviate the limitation, learning-based strategy which bases on variable interaction characteristics of the objective function was proposed. CCVIL [17], delta grouping [18], LINC-R [19] and CCEA-AVP [20] all belong to learning-based strategy and their performance are much better than the former two strategies. But a common drawback for these algorithms is their low grouping accuracy.

DG is another learning-based algorithm which was proposed by Omidvar *et al* [9]. It focuses on detecting additive separability which is the most common type of partial separability. Compared with the former algorithms, the performance of DG is much better. In DG, interaction between decision variables is identified according to the following rule:

**Theorem 1:** Let $f(\vec{x})$ be an additively separable function. $\forall a$, $b_1 \neq b_2, \delta \in \mathbb{R}, \delta \neq 0$, if the following condition holds:

$$\Delta_{\delta, x_p}[f](\vec{x})|_{x_p=a, x_q=b_1} \neq \Delta_{\delta, x_p}[f](\vec{x})|_{x_p=a, x_q=b_2} \quad (1)$$

where

$$\Delta_{\delta, x_p}[f](\vec{x}) = f(\ldots, x_p + \delta, \ldots) - f(\ldots, x_p, \ldots) \quad (2)$$

*refers to the forward difference of f with respect to variable $x_p$ with interval $\delta$.*

According to Theorem 1, when change in $f$ caused by adding a perturbation to $x_p$ varies for different value of $x_q$, then $x_p$ interacts with $x_q$. But when in practice, computational roundoff error will be generated for the reason of limited precision of floating-point numbers, thus a threshold parameter ($\varepsilon$) need to be set to compensate for the roundoff error. Specifically, when we denote the left side of Equation (1) as $\Delta_1$ and the right side as $\Delta_2$, if the following condition holds:

$$\tau = |\Delta_1 - \Delta_2| > \varepsilon \tag{3}$$

then $x_p$ and $x_q$ are nonseparable.

Despite its success, DG still has some drawbacks. To be specific, it cannot detect indirect interactions and consumes lots of FEs on some functions. Besides, its decomposition accuracy is also limited by the threshold setting method. After DG was proposed, many algorithms were developed to improved it, including global DG [11], extend DG [10], FII [12], DG2 [13] and VGDA [14]. The first two drawbacks have been remedied very well, but for threshold setting, some work still need to be done to improve it.

## 2.2 Threshold Setting Methods

The threshold parameter has a great influence on the decomposition accuracy of DG-based algorithms, there are three methods to set it up to now, including fixed threshold (FT) [9], function space based threshold (FST) [11] and computational roundoff error based threshold (CRET) [13]. In this part, we will review them in sequence.

FT is the simplest method which is adopted by DG, XDG and FII as their threshold setting method. In FT, a fixed value, which is usually $10^{-1}$ or $10^{-3}$, is selected as the threshold parameter for all functions being decomposed. However, for the reason that roundoff error is related to the objective function, a fixed threshold parameter cannot be suitable for all kinds of functions. Therefore, the robustness of FT is very poor.

FST is another threshold setting method which was proposed in GDG. In FST, the threshold is set according to the magnitude of the objective function value. Specifically, the threshold is generated based on the following equation:

$$\varepsilon = \alpha \cdot \min\{|f(x_1)|, |f(x_2)|, \ldots, |f(x_k)|\} \tag{4}$$

where $x_1, x_2, \ldots, x_k$ are $k$ randomly sample points in solution space, and $\alpha$ is the control parameter. Compared with FT, the performance of FST has been improved a lot. But according to [13], when FST is employed on the functions with imbalanced contribution weights of different sub-problems, the final decomposition results are unsatisfactory.

Different from the former two methods, Omidvar *et al* [13] proposed another threshold setting method called CRET in DG2. It is a parameter-free method and the threshold is generated from the corresponding roundoff error. For CRET, it first calculates the greatest lower bound ($e_{inf}$) and the least upper bound ($e_{sup}$) for the roundoff error generated in Equation (3) for each pair of variables. For the pair of variables, $x_p$ and $x_q$, their $e_{inf}$ and $e_{sup}$ can be obtained as follows:

$$e_{inf} = \gamma_2 \cdot \max\{|f(x_1) + f(x_4)|, \ |f(x_2) + f(x_3)|\} \tag{5}$$

$$e_{sup} = \gamma_{\sqrt{n}} \cdot \max\{f(x_1), f(x_2), f(x_3), f(x_4)\} \tag{6}$$

where $x_1 = (\ldots, x_p, \ldots, x_q, \ldots)$, $x_2 = (\ldots, x_p', \ldots, x_q, \ldots)$, $x_3 = (\ldots, x_p, \ldots, x_q', \ldots)$ and $x_4 = (\ldots, x_p', \ldots, x_q', \ldots)$ are four points in solution space generated during the calculation of the quantity $\tau$ of $x_p$ and $x_q$, $\gamma_2$ and $\gamma_{\sqrt{n}}$ are two small constants. Then for each pair of variables, a corresponding threshold will be estimated based on their $e_{inf}$ and $e_{sup}$.

## 3 GLOBAL INFORMATION BASED ADAPTIVE THRESHOLD SETTING ALGORITHM

In this section, the proposed threshold setting method GIAT will be introduced. The new indicator for interaction between variables will be defined firstly, and the process of threshold setting based on the indicator will be given next. The overall GIAT will be shown in the final.

### 3.1 Indicator for Interaction between Variables

Theorem 1 is the foundation of most DG-based algorithms and the quantity $\tau$ in Equation (3) is the key to determine whether two variables are separable. For two separable variables, their $\tau$ value is just the roundoff error, so $\tau$ is very sensitive to the roundoff error, which make it difficult to find out a suitable threshold parameter for some problems. Therefore, the sensitivity of $\tau$ on the computational roundoff error needs to be reduced.

By taking the corresponding greatest lower bound ($e_{inf}$) for roundoff error of $\tau$ into account, each $\tau$ value can be transformed as follow to reduce its sensitivity to the roundoff error:

$$\tau' = (\tau - e_{inf}) \cdot f_s(\tau - e_{inf}) \tag{7}$$

where $e_{inf}$ is the corresponding greatest lower bound for the roundoff error of $\tau$, and $f_s(\cdot)$ is a unit step function.

The role of Equation (7) is to significantly reduce the quantity $\tau$ for separable variables as it is just the roundoff error while the other $\tau$ values are almost unchanged. As a result, the $\tau'$ for separable variables will be either zero or extremely tiny, so it is much less sensitive to roundoff error.

Besides, the $\tau$ for nonseparable variables is proportional to the contribution weight of subcomponent which the two nonseparable variables belong to, such as Example 1:

EXAMPLE 1. *In the objective function:* $f(\vec{x}) = \omega_1 \cdot (x_1 - x_2)^2 + \omega_2 \cdot (x_3 - x_4)^2$, $\vec{x} \in [-1, 1]^4$, *if we employ the perturbation standard that all the variables are initialized as the lower bound, and then perturbed to the upper bound, the $\tau$ for $x_1$ and $x_2$ is $8\omega_1$, and that for $x_3$ and $x_4$ is $8\omega_2$.*

Therefore, $\tau$ is also very sensitive to the magnitude of the contribution weight of subcomponent. For the function with unbalanced contribution weights of different subcomponents, the suitable threshold is difficult to find out. To alleviate the limitation, the $\tau'$ should be transformed further as follow:

$$\tau'' = \frac{\tau'}{\max(|\Delta_1|, |\Delta_2|)} \tag{8}$$

where $\Delta_1$ and $\Delta_2$ are the left side and right side of Equation (1), respectively, which are also proportional to the corresponding contribution weight.

In summary, for $x_p$ and $x_q$, their interaction indicator ($\zeta$) can be defined as follow:

$$\zeta = \frac{(|\Delta_1 - \Delta_2| - e_{\inf}) \cdot f_s(|\Delta_1 - \Delta_2| - e_{\inf})}{\max(|\Delta_1|, |\Delta_2|)} \quad (9)$$

where $\Delta_1$ and $\Delta_2$ are the left side and right side of Equation (1) for $x_p$ and $x_q$, respectively, and $e_{\inf}$ is the corresponding greatest lower bound for the roundoff error generated during the calculation of $\tau$ for $x_p$ and $x_q$, $f_s(\cdot)$ is a unit step function.

## 3.2 Threshold Setting Based on Global Information

It has been shown that the interaction indicator is much more sensitive to the interaction between variables and less sensitive to computational roundoff error and contribution weight of different subcomponent. Therefore, if there is a big difference between two interaction indicators, the smaller one must be the indicator for two separable variables and the bigger one must be the indicator for two nonseparable variables. As a conclusion, for an array **Z** which stores the interaction indicators for each pair of variables, if we sort the elements of **Z** in ascending order, the two adjacent indicators with the greatest difference are the biggest interaction indicator for separable variables and the smallest interaction indicator for nonseparable variables, respectively. Then any values in the range of the two critical interaction indicators can be set as the threshold to distinguish the interaction indicator for separable variables and that for nonseparable variables.

It is necessary to note that the above process is just designed for partially separable function. So before putting it into practice, we should exclude the fully separable and nonseparable functions. Based on [13], if the quantity $\tau$ for two variables is larger than its corresponding greatest lower bound ($e_{\inf}$), the two variables are separable. And if the $\tau$ is less than its corresponding least upper bound ($e_{\sup}$), the two variables are nonseparable. Therefore, the corresponding $e_{\inf}$ and $e_{\sup}$ for each $\tau$ values can help us nearly reach the aim of excluding fully separable or nonseparable functions. Specifically, for a problem where all $\tau$ values are less than their corresponding $e_{\inf}$, it is fully separable. On the contrary, if all the $\tau$ values are bigger than their corresponding $e_{\sup}$, the problem is fully nonseparable.

Fig. 1 shows the framework of the threshold parameter setting process. For the fully separable or nonseparable functions, their threshold can be set easily. However, how to evaluate the difference between each two adjacent indicators in **Z** is an important problem. To eliminate the influence of magnitude of different indicators, we adopt the quotient of two indicators as the measure to evaluate their difference. Specifically, if the quotient of two interaction indicators is larger than that of another two, the difference of the former two is greater.

*Remark*: There is a problem in the above measure that how to evaluate the difference between the last zero interaction indicator and the first non-zero interaction indicator in **Z**. For the partially separable problems where all $\tau$ values are out of the range of their corresponding $e_{\inf}$ and $e_{\sup}$, the interaction indicators for each pair of separable variables are zero based on Equation (9), so the last zero interaction indicator and the first non-zero interaction indicator in **Z** are the two critical interaction indicators and we can directly set the threshold parameter as zero. As for other problems, both of the two values are the interaction indicators for two separable variables, so we can set their difference as zero. In this way, the measure can work correctly.

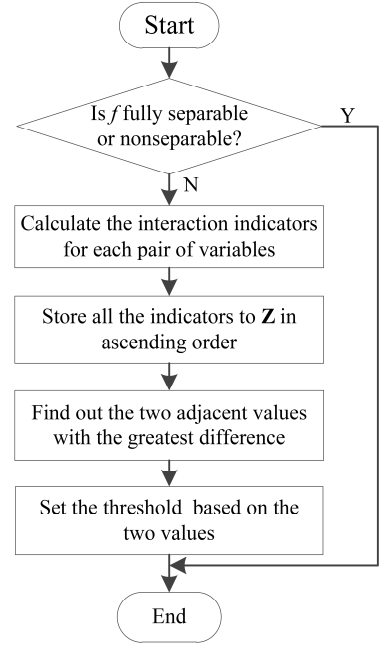

**Figure 1: The framework of threshold parameter setting process.**

## 3.3 Overall Implementation

Algorithm 1 shows the overall GIAT algorithm. The inputs of GIAT for an $n$-dimensional function $f$ include the interaction structure matrix $\mathbf{\Gamma}_{n \times n}$ which contains the $\tau$ for all pair of variables, the matrix $\mathbf{E}_{n \times n}^{\inf}$ which contains the corresponding $e_{\inf}$ of each $\tau$ in $\mathbf{\Gamma}_{n \times n}$, the matrix $\mathbf{E}_{n \times n}^{\sup}$ which contains the corresponding $e_{\sup}$ of each $\tau$ in $\mathbf{\Gamma}_{n \times n}$, and the matrix $\mathbf{D}_{n \times n}$ which contains the corresponding $\max(|\Delta_1|, |\Delta_2|)$ of each $\tau$ in $\mathbf{\Gamma}_{n \times n}$.

There are three stages in GIAT. The first stage is to exclude the fully separable or nonseparable functions (Lines 01-05). The second stage is to calculate the interaction indicators for each pair of variables based on Equation (9) and save them into the array **Z** in ascending order (Lines 06-09). The third stage is to find out the two adjacent values with the biggest difference and estimate the threshold based on them (Lines 10-24). For the third stage, the problem should be classified into two categories as mentioned above for the reason of difference measure adopted. Besides, we directly adopt the smaller indicator as the threshold parameter.

Last but not least, while the obtained threshold parameter is employed in the decomposition process, the interaction structure matrix should become the one which contains the interaction indicators for each pair of variables.

**Algorithm 1:** GIAT

Require: $\mathbf{\Gamma}_{n \times n}$, $\mathbf{E}_{n \times n}^{\inf}$, $\mathbf{E}_{n \times n}^{\sup}$, $\mathbf{D}_{n \times n}$
01 **if** $\mathbf{\Gamma}_{n \times n} < \mathbf{E}_{n \times n}^{\inf}$
02     $f$ is fully separable and output $\varepsilon$ = inf;
03 **else if** $\mathbf{\Gamma}_{n \times n} > \mathbf{E}_{n \times n}^{\sup}$
04     $f$ is fully nonseparable and output $\varepsilon$ = 0;
05 Calculate the $\zeta$ for each pair of variables based on Eq. (9);
06 Store all $\zeta$ values in array $\mathbf{Z}$ and set the negatives to 0;
07 Sort $\mathbf{Z}$ in ascending order;
08 Generate array $\mathbf{V}$ to store the difference value of each two adjacent values in $\mathbf{Z}$;
09 **if** isempty ( $\mathbf{E}_{n \times n}^{\inf} < \mathbf{\Gamma} < \mathbf{E}_{n \times n}^{\sup}$ )
10     Output $\varepsilon$ = 0;
11 **else**
12     **for** $i$ = 2 : length($\mathbf{Z}$)
13       **if** $\mathbf{Z}(i-1) == 0$
14          $\mathbf{V}(i) = 0$;
15       **else**
16          $\mathbf{V}(i) = \mathbf{Z}(i) / \mathbf{Z}(i-1)$
17     $j = \arg\max_j (\mathbf{V}(j))$
18     Output $\varepsilon = \mathbf{Z}(j-1)$;

## 4 EXPERIMENT STUDIES

In this section, detailed numerical experiments are conducted to investigate the efficacy of GIAT. Specifically, two experiments will be conducted. The first experiment investigates the effectiveness of proposed interaction indicator for two variables. The second experiment evaluates the overall performance of GIAT by the comparative analysis on it and other threshold setting methods.

### 4.1 Benchmark Functions in the CEC2010 and CEC2013 Suites

The experiments will be conducted on the functions in CEC2010 and CEC2013 benchmark suites, a brief introduction of the two benchmark suites is given below. There are totally 20 and 15 functions in CEC2010 and CEC2013 suites, respectively. The 20 functions in CEC2010 benchmark suite can be classified into 5 categories [15], and the 15 functions in CEC2013 benchmark suite can be classified into 4 categories [16]. The detailed categories for the two sets of functions are shown in Table 1.

It is necessary to note that the overlapping functions $f_{12}$ to $f_{14}$ in CEC2013 are regarded as fully nonseparable functions. Compared with CEC2010 functions, much more complicated transformations are introduced into CEC2013 functions, which make them more difficult to decompose. Besides, the dimensions and the contribution weights of different sub-problems are very imbalanced in CEC2013 functions.

### 4.2 Effectiveness of Interaction Indicator

**Table 1: The Categories of Functions in CEC2010 and CEC2013 Benchmark Suites. NPS means nonseparable subcomponents.**

| Suite | F | Categories |
|---|---|---|
| CEC 2010 | $f_1$-$f_3$ | Fully separable functions |
| | $f_4$-$f_8$ | Partially separable functions with 1 NPS |
| | $f_9$-$f_{13}$ | Partially separable functions with 10 NPS |
| | $f_{14}$-$f_{18}$ | Partially separable functions with 20 NPS |
| | $f_{19}$-$f_{20}$ | Fully nonseparable functions |
| CEC 2013 | $f_1$-$f_3$ | Fully separable functions |
| | $f_4$-$f_7$ | Partially separable functions with 7 NPS |
| | $f_8$-$f_{11}$ | Partially separable functions with 20 NPS |
| | $f_{13}$-$f_{15}$ | Fully nonseparable functions |

In this section, we will study the effectiveness of the proposed interaction indicator on some typical functions in the two benchmark suites. When setting the threshold parameter, the array $\mathbf{Z}$ stores the interaction indicators for all pair of variables in ascending order. Therefore, we can investigate the difference information about each two adjacent values in $\mathbf{Z}$ to achieve our aim.

Fig. 2 presents the distribution graphs for the difference between each two adjacent values of the final $\mathbf{Z}$ on $f_4$, $f_{15}$ in CEC2010 and $f_7$, $f_{11}$ in CEC2013, which have been all correctly decomposed. Fig.1(a) and Fig.1(b) show the difference distribution graphs of $f_4$ and $f_{15}$, respectively. From the two graphs, we can find that there is only one point whose value is much greater than any other points and it corresponds to the difference between the biggest interaction indicator for separable variables and the smallest interaction indicator for nonseparable variables. For the two functions $f_4$ and $f_{15}$, the dimensions and the contribution weights of different sub-problems are consistent, so the influencing factors for the interaction indicator are the roundoff error and the interaction between variables. From the results we can find that the interaction indicator is only sensitive to the latter.

The excellent performance of interaction indicator is further verified on $f_7$ and $f_{11}$ in CEC2013. For these two functions, the contribution weights of different subcomponents have a great difference, so the influencing factors for the interaction indicator include the roundoff error, the contribution weights of different subcomponents and the interaction between variables. From the results in Fig. 1(c) and Fig. 1(d), it can be observed that only the difference between the biggest interaction indicator for separable variables and the smallest interaction indicator for nonseparable variables is very large and the differences for any other two adjacent elements in $\mathbf{Z}$ are much smaller. In conclusion, the interaction indicator is no longer sensitive to the roundoff error and the magnitude of the contribution weights of different subcomponents, and the only influencing factor is the interaction between variables.

The four graphs in Fig. 2 verified the prior performance of interaction indicator. It is very sensitive to the interaction between variables and compared with the $\tau$ in DG, it is much less sensitive to the roundoff error and the contribution weight of subcomponent.

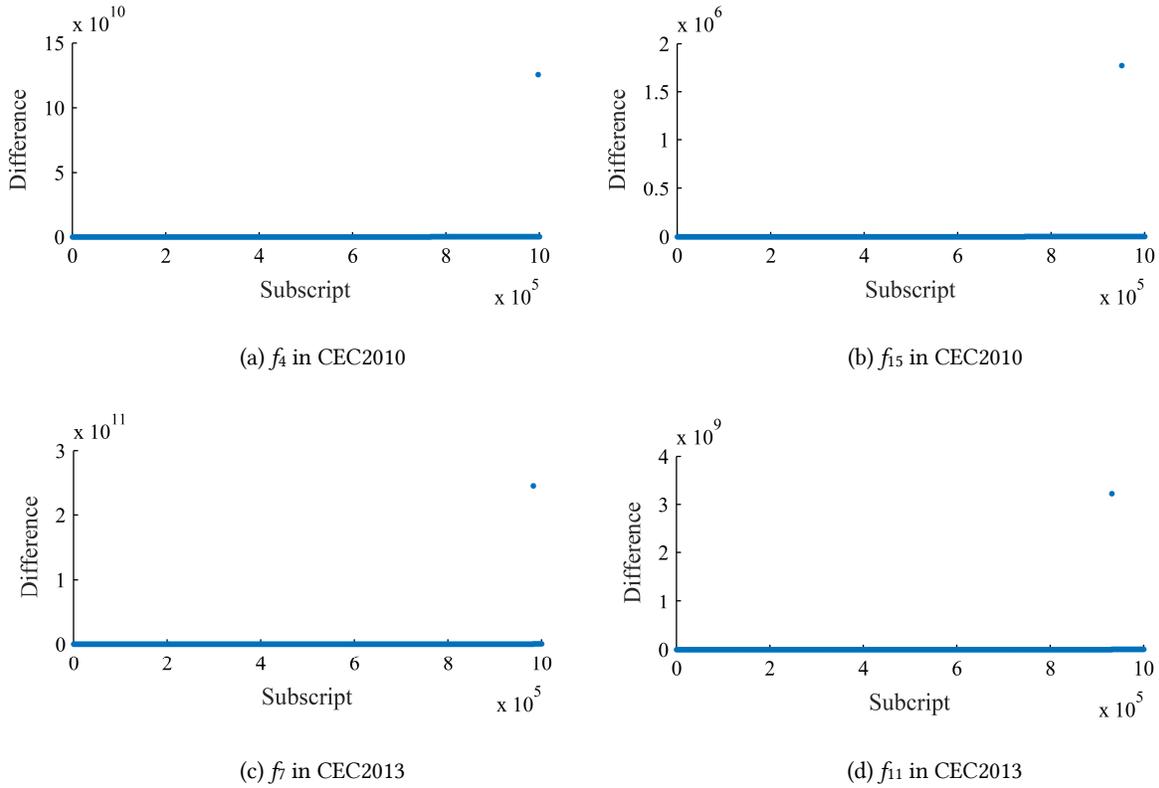

**Figure 2: The difference distribution graphs of the final Z on four functions.**

## 4.3 Comparison between GIAT and other Threshold Setting Methods

To have a comprehensive knowledge of the effectiveness of GIAT, we compared it with the other three threshold parameter setting methods, including FT, FST and CRET, all of which have been introduced in Section 2. To ensure the fairness of comparison, the parameters in the comparative algorithms are set as their original papers. Specifically, the threshold in FT is set as $10^{-3}$, and the control parameter in FST is set as $10^{-10}$. Besides, all the four methods are combined with the decomposition framework of GDG where a raw interaction structure matrix is built. For convenience, the combined decomposition algorithms will be denoted as the corresponding threshold setting methods.

Table 2 summarizes the final decomposition results obtained by the four algorithms on each function in CEC2010 and CEC2013 benchmark suites. The results of each method are separated by '/' and if all variables are correctly identified and grouped, its accuracy is 1, otherwise it is 0. Besides, the double lines make a clear distinction for the different classes of functions according to [15] and [16].

From the results on the functions in CEC2010 benchmark suite, it can be observed that GIAT performs very well. In summary, GIAT can correctly decompose 18 functions in all 20 functions. As for the other three algorithms, FST can also obtain the correct decomposition results on 18 functions, FT and CRET can correctly decompose 15 and 17 functions, respectively. Therefore, GIAT and FST tied for the first among the four algorithms in CEC2010 benchmark suite, which shows the excellent performance of GIAT.

The decomposition results on CEC2013 benchmark suite further demonstrate the excellent performance of GIAT. For the reason that the dimensions and the contribution weights of different sub-problems are very imbalanced, the performance of FT and FST degenerate a lot, they can only correctly decompose 9 and 8 in all 15 functions, respectively. Even so, both CRET and GIAT can still obtain ideal decomposition accuracy as they can correctly decompose 12 functions.

When concerning the functions which are improperly decomposed by GIAT, all of their basic functions are the *Ackley* functions except $f_8$ in CEC2013. It is known that *Ackley* function is not additively separable, and the interaction detection method in DG is not suitable for it. Even so, CVAT can correctly decompose some *Ackley* functions, such as $f_6$ in CEC2010. As for $f_8$ in CEC2013, most variables are correctly grouped and only a few of them are improperly grouped.

In summary, GIAT ranks first on both CEC2010 and CEC2013 suites, which shows its high efficiency and strong robustness. As a conclusion, GIAT outperforms the existing threshold setting methods and it can be seen as an excellent threshold parameter setting method.

Table 2: The decomposition results obtained by GDG framework with different threshold setting method. Sep-Vars and Nonsep-Vars mean the number of separable variables and the number of nonseparable variables, respectively.

| | F | Sep-Vars | Nonsep-Vars | Nonsep-Groups | FT ($\varepsilon = 10^{-3}$) / FST / CRET / GIAT | | | |
|---|---|---|---|---|---|---|---|---|
| | | | | | Captured Sep-Vars | Captured Nonsep-Vars | Formed Nonsep Subcomponents | Accuracy |
| CEC2010 Benchmark Suite | $f_1$ | 1000 | 0 | 0 | 1000 / 1000 / 1000 / 1000 | 0 / 0 / 0 / 0 | 0 / 0 / 0 / 0 | 1 / 1 / 1 / 1 |
| | $f_2$ | 1000 | 0 | 0 | 1000 / 1000 / 1000 / 1000 | 0 / 0 / 0 / 0 | 0 / 0 / 0 / 0 | 1 / 1 / 1 / 1 |
| | $f_3$ | 1000 | 0 | 0 | 1000 / 0 / 0 / 0 | 0 / 1000 / 1000 / 1000 | 0 / 1 / 1 / 1 | 1 / 0 / 0 / 0 |
| | $f_4$ | 950 | 50 | 1 | 0 / 950 / 950 / 950 | 1000 / 50 / 50 / 50 | 1 / 1 / 1 / 1 | 0 / 1 / 1 / 1 |
| | $f_5$ | 950 | 50 | 1 | 950 / 950 / 950 / 950 | 50 / 50 / 50 / 50 | 1 / 1 / 1 / 1 | 1 / 1 / 1 / 1 |
| | $f_6$ | 950 | 50 | 1 | 950 / 950 / 851 / 950 | 50 / 50 / 149 / 50 | 1 / 1 / 2 / 1 | 1 / 1 / 0 / 1 |
| | $f_7$ | 950 | 50 | 1 | 0 / 950 / 950 / 950 | 1000 / 50 / 50 / 50 | 1 / 1 / 1 / 1 | 0 / 1 / 1 / 1 |
| | $f_8$ | 950 | 50 | 1 | 0 / 950 / 950 / 950 | 1000 / 50 / 50 / 50 | 2 / 1 / 1 / 1 | 0 / 1 / 1 / 1 |
| | $f_9$ | 500 | 500 | 10 | 500 / 500 / 500 / 500 | 500 / 500 / 500 / 500 | 10 / 10 / 10 / 10 | 1 / 1 / 1 / 1 |
| | $f_{10}$ | 500 | 500 | 10 | 500 / 500 / 500 / 500 | 500 / 500 / 500 / 500 | 10 / 10 / 10 / 10 | 1 / 1 / 1 / 1 |
| | $f_{11}$ | 500 | 500 | 10 | 500 / 0 / 0 / 0 | 500 / 1000 / 1000 / 1000 | 10 / 11 / 11 / 11 | 1 / 0 / 0 / 0 |
| | $f_{12}$ | 500 | 500 | 10 | 500 / 500 / 500 / 500 | 500 / 500 / 500 / 500 | 10 / 10 / 10 / 10 | 1 / 1 / 1 / 1 |
| | $f_{13}$ | 500 | 500 | 10 | 4 / 500 / 500 / 500 | 996 / 500 / 500 / 500 | 2 / 10 / 10 / 10 | 0 / 1 / 1 / 1 |
| | $f_{14}$ | 0 | 1000 | 20 | 0 / 0 / 0 / 0 | 1000 / 1000 / 1000 / 1000 | 20 / 20 / 20 / 20 | 1 / 1 / 1 / 1 |
| | $f_{15}$ | 0 | 1000 | 20 | 0 / 0 / 0 / 0 | 1000 / 1000 / 1000 / 1000 | 20 / 20 / 20 / 20 | 1 / 1 / 1 / 1 |
| | $f_{16}$ | 0 | 1000 | 20 | 0 / 0 / 0 / 0 | 1000 / 1000 / 1000 / 1000 | 20 / 20 / 20 / 20 | 1 / 1 / 1 / 1 |
| | $f_{17}$ | 0 | 1000 | 20 | 0 / 0 / 0 / 0 | 1000 / 1000 / 1000 / 1000 | 20 / 20 / 20 / 20 | 1 / 1 / 1 / 1 |
| | $f_{18}$ | 0 | 1000 | 20 | 0 / 0 / 0 / 0 | 1000 / 1000 / 1000 / 1000 | 1 / 20 / 20 / 20 | 0 / 1 / 1 / 1 |
| | $f_{19}$ | 0 | 1000 | 1 | 0 / 0 / 0 / 0 | 1000 / 1000 / 1000 / 1000 | 1 / 20 / 20 / 20 | 1 / 1 / 1 / 1 |
| | $f_{20}$ | 0 | 1000 | 1 | 0 / 0 / 0 / 0 | 1000 / 1000 / 1000 / 1000 | 1 / 20 / 20 / 20 | 1 / 1 / 1 / 1 |
| | Summary | | | | — | — | — | 15 / **18** / 17 / **18** |
| CEC2013 Benchmark Suite | $f_1$ | 1000 | 0 | 0 | 1000 / 1000 / 1000 / 1000 | 0 / 0 / 0 / 0 | 0 / 0 / 0 / 0 | 1 / 1 / 1 / 1 |
| | $f_2$ | 1000 | 0 | 0 | 1000 / 1000 / 1000 / 1000 | 0 / 0 / 0 / 0 | 0 / 0 / 0 / 0 | 1 / 1 / 1 / 1 |
| | $f_3$ | 1000 | 0 | 0 | 1000 / 0 / 0 / 0 | 0 / 1000 / 1000 / 1000 | 0 / 1 / 1 / 1 | 1 / 0 / 0 / 0 |
| | $f_4$ | 700 | 300 | 7 | 0 / 700 / 700 / 700 | 1000 / 300 / 300 / 300 | 1 / 7 / 7 / 7 | 0 / 1 / 1 / 1 |
| | $f_5$ | 700 | 300 | 7 | 700 / 770 / 700 / 700 | 300 / 230 / 300 / 300 | 7 / 7 / 7 / 7 | 1 / 0 / 1 / 1 |
| | $f_6$ | 700 | 300 | 7 | 750 / 750 / 0 / 0 | 250 / 250 / 1000 / 1000 | 6 / 5 / 1 / 8 | 0 / 0 / 0 / 0 |
| | $f_7$ | 700 | 300 | 7 | 0 / 700 / 700 / 700 | 1000 / 300 / 300 / 300 | 1 / 7 / 7 / 7 | 0 / 1 / 1 / 1 |
| | $f_8$ | 0 | 1000 | 20 | 0 / 398 / 200 / 97 | 1000 / 602 / 800 / 903 | 1 / 19 / 18 / 25 | 0 / 0 / 0 / 0 |
| | $f_9$ | 0 | 1000 | 20 | 0 / 22 / 0 / 0 | 1000 / 978 / 1000 / 1000 | 20 / 20 / 20 / 20 | 1 / 0 / 1 / 1 |
| | $f_{10}$ | 0 | 1000 | 20 | 150 / 150 / 0 / 0 | 850 / 850 / 1000 / 1000 | 17 / 17 / 20 / 20 | 0 / 0 / 1 / 1 |
| | $f_{11}$ | 0 | 1000 | 20 | 0 / 0 / 0 / 0 | 1000 / 1000 / 1000 / 1000 | 1 / 1 / 20 / 20 | 0 / 0 / 1 / 1 |
| | $f_{12}$ | 0 | 1000 | 1 | 0 / 0 / 0 / 0 | 1000 / 1000 / 1000 / 1000 | 1 / 1 / 1 / 1 | 1 / 1 / 1 / 1 |
| | $f_{13}$ | 0 | 1000 | 1 | 0 / 0 / 0 / 0 | 905 / 905 / 905 / 905 | 1 / 1 / 1 / 1 | 1 / 1 / 1 / 1 |
| | $f_{14}$ | 0 | 1000 | 1 | 0 / 0 / 0 / 0 | 905 / 905 / 905 / 905 | 1 / 1 / 1 / 1 | 1 / 1 / 1 / 1 |
| | $f_{15}$ | 0 | 1000 | 1 | 0 / 0 / 0 / 0 | 1000 / 1000 / 1000 / 905 | 1 / 1 / 1 / 1 | 1 / 1 / 1 / 1 |
| | Summary | | | | — | — | — | 9 / 8 / **12** / **12** |

## 5 CONCLUSIONS

In this paper, an adaptive threshold parameter setting method named GIAT for DG-based algorithms was developed. Different from the other algorithms, the threshold parameter in GIAT is estimated based on the global information. Specifically, a new indicator is developed in GIAT which is very sensitive to the interaction between variables and less sensitive to the roundoff error and the magnitude of the contribution weights of subcomponents. Only between the indicator for two separable variables and the indicator for two nonseparable variables, there will be great difference. Therefore, by sorting the indicators for each pair of variables in ascending order, the two indicators with the greatest difference are the biggest value for two separable variables and the smallest value for two nonseparable variables, respectively. And a threshold can be easily set based on the two values. The superior and robust performance of GIAT is obtained without setting any parameter, which makes it more applicable for different problems.

Our future work will focus on improving GIAT so that it can correctly decompose fully separable or nonseparable functions rather than adopting the present rough method. In this way, GIAT can be combined with some high-efficiency decomposition algorithms to reduce the cost of fitness evaluations. Then it can be well integrated into the CC framework.

## ACKNOWLEDGMENTS

This work was supported in part by the National Natural Science Foundation of China under Grant 61105126 and in part by the Postdoctoral Science Foundation of China under Grants 2014M560784 and 2016T90922.